%% file: neurips_2024.tex
\documentclass{article}

\usepackage[numbers]{natbib}
\usepackage[final]{neurips_2024}

\usepackage[utf8]{inputenc} 
\usepackage[T1]{fontenc}    
\usepackage{hyperref}       
\usepackage{url}            
\usepackage{booktabs}       
\usepackage{amsfonts}       
\usepackage{nicefrac}       
\usepackage{microtype}      
\usepackage{xcolor}         
\usepackage{tabularx}       
\usepackage{graphicx}
\usepackage{lipsum}
\usepackage{marvosym}
\usepackage{xspace}
\usepackage{blindtext}
\usepackage{amsmath}
\usepackage{algorithm}
\usepackage{algpseudocode}
\usepackage{wrapfig}
\usepackage{enumitem}
\usepackage{subfigure}

\hypersetup{colorlinks,urlcolor=black,citecolor=cyan}

\makeatletter
\DeclareRobustCommand\onedot{\futurelet\@let@token\@onedot}
\def\@onedot{\ifx\@let@token.\else.\null\fi\xspace}

\def\eg{\emph{e.g}\onedot}

\def\etc{\emph{etc}\onedot}

\makeatother

\def\shortname{\mbox{HunyuanWorld 1.0}\xspace}

\definecolor{citecolor}{rgb}{0.21,0.49,0.74}
\definecolor{linkcolor}{HTML}{ED1C24}
\definecolor{graycolor}{rgb}{0.95,0.95,0.95}

\usepackage[capitalize]{cleveref}
\crefname{section}{Sec.}{Secs.}
\crefname{table}{Tab.}{Tabs.}
\crefname{figure}{Fig.}{Figs.}

\newcommand{\figref}[1]{Fig.~\ref{#1}}

\title{
\shortname: Generating Immersive,  Explorable, and Interactive 3D Worlds from Words or Pixels
}

\author{
\textbf{Tencent Hunyuan*}
}
\begin{document}

\maketitle

\input{sections/abstract}

\begin{figure}[h]
\centering
\includegraphics[width=0.95\textwidth]{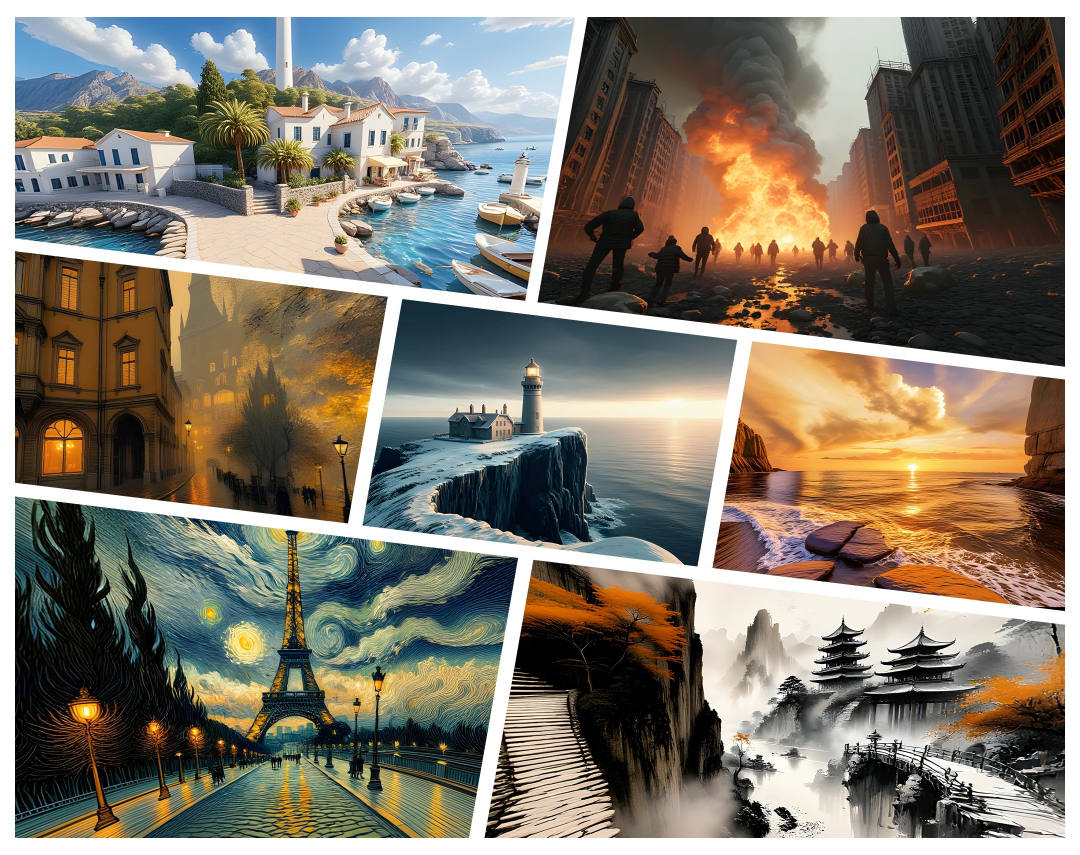}
\end{figure}

\makeatletter
\def\@makefnmark{}
\makeatother
\footnotetext{$*$ HunyuanWorld 1.0 team contributors are listed in the end of report.}

\clearpage
\begin{figure}[t]
\centering
\includegraphics[width=\textwidth]{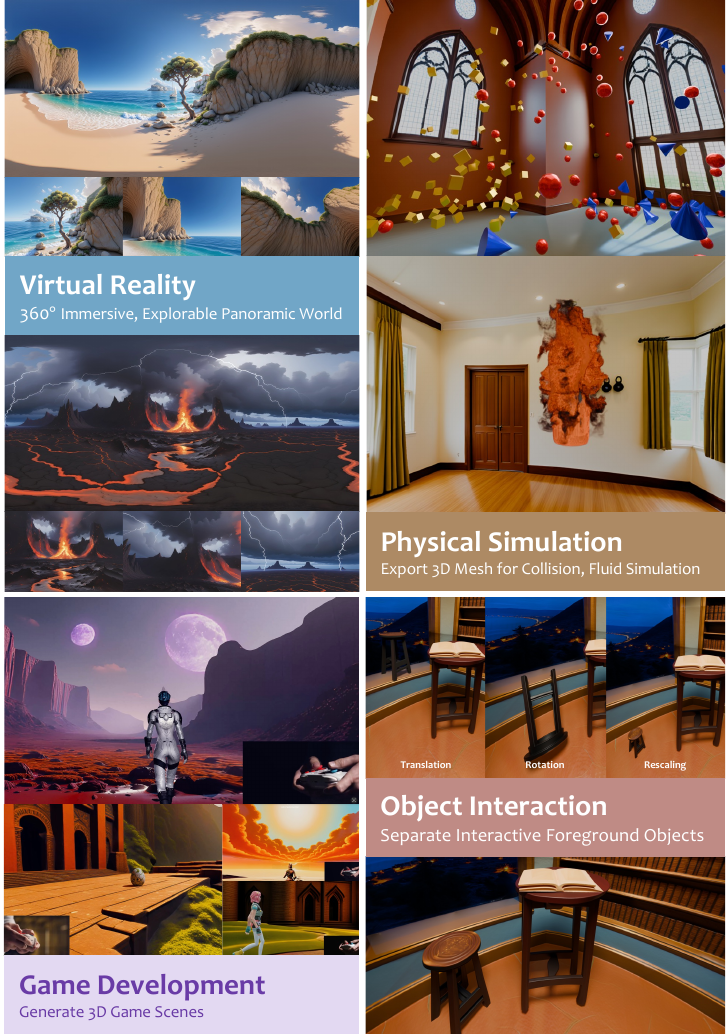}
\caption{An overview of \shortname applications. }
\label{fig:application}
\end{figure}
\clearpage

\input{sections/introduction}

\input{sections/method}
\input{sections/experiments}
 \input{sections/related-work}

\input{sections/conclusion}

\clearpage

{\small
\bibliographystyle{plain}
\bibliography{references}
}

\end{document}

%% file: sections/abstract.tex
\begin{abstract}
Creating immersive and playable 3D worlds from texts or images remains a fundamental challenge in computer vision and graphics. Existing world generation approaches typically fall into two categories: video-based methods that offer rich diversity but lack 3D consistency and rendering efficiency, and 3D-based methods that provide geometric consistency but struggle with limited training data and memory-inefficient representations. To address these limitations, we present \shortname, a novel framework that combines the best of both worlds for generating immersive, explorable, and interactive 3D scenes from text and image conditions. Our approach features three key advantages: 1) 360° immersive experiences via panoramic world proxies; 2) mesh export capabilities for seamless compatibility with existing computer graphics pipelines; 3) disentangled object representations for augmented interactivity. The core of our framework is a semantically layered 3D mesh representation that leverages panoramic images as 360° world proxies for semantic-aware world decomposition and reconstruction, enabling the generation of diverse 3D worlds.   Extensive experiments demonstrate that our method achieves state-of-the-art performance in generating coherent, explorable, and interactive 3D worlds while enabling versatile applications in virtual reality, physical simulation, game development, and interactive content creation.
\end{abstract}

%% file: sections/introduction.tex
\section{Introduction}

\begin{quote}
\raggedleft \emph{``To see a World in a Grain of Sand, and a Heaven in a Wild Flower'' } \\
\raggedleft \emph{--- William Blake}
\end{quote}

World models have emerged as a fundamental paradigm for understanding and generating complex 3D environments, with applications spanning virtual reality, autonomous driving, robotics, and video gaming. The ability to create immersive, explorable, and interactive 3D worlds from natural language descriptions or visual inputs represents a crucial milestone toward democratizing 3D content creation and enabling new forms of human-computer interaction.

Recently, world generation has achieved remarkable progress, and the main solutions can be divided into two categories: video-based world generation methods and 3D-based world generation methods. Video-based methods leverage the inherent world knowledge of video diffusion models~\cite{kong2024hunyuanvideo,yang2024cogvideox,wan2025wan, singer2022make,teng2025magi,lin2024open,blattmann2023stable} to understand the temporal and spatial relationships of the generated world. The rich training data available for video models enables them to capture complex real-world dynamics for generating visually compelling results across diverse scenarios. Some works further incorporate 3D constraints such as camera trajectories~\cite{he2025cameractrl,agarwal2025cosmos,alhaija2025cosmos,sun2024dimensionx} or explicit 3D scene point clouds~\cite{huang2025voyager,wu2025video} to spatially control the generated video sequences and produce plausibly 3D consistent video worlds. 

However, video-based approaches face several fundamental limitations that constrain their practical performance. First, they inherently lack true 3D consistency due to their underlying 2D frame-based representation. This leads to temporal inconsistencies, particularly when generating long-range video scenes where accumulated errors result in severe content drift and incoherence. Second, the rendering costs of video-based methods are prohibitive, as each frame should be generated sequentially. Third, the frame-based video format is fundamentally incompatible with existing computer graphics pipelines, making it hard to be incorporated into game engines, VR applications, and other interactive systems.

In contrast, 3D-based world generation methods directly model geometric structures and offer superior compatibility with current computer graphics pipelines. These approaches provide inherent 3D consistency with efficient real-time rendering. Despite rapid advances in object-level 3D generation~\cite{poole2022dreamfusion, wang2023rodin,Tang_2023_ICCV,  wang2024phidias, hong2023lrm,openlrm,tang2024lgm,xiang2024structured,yang2024hunyuan3d, hunyuan3d22025tencent, lai2025hunyuan3d25highfidelity3d}, world-level 3D generation remains significantly underexplored. 
Although some recent works~\cite{chung2023luciddreamer,yu2024wonderjourney,yu2025wonderworld,yang2024layerpano3d, worldgen} have demonstrated the potential of generating multi-view consistent 3D scenes from text descriptions, world-level 3D synthesis remains constrained by several critical challenges. The primary limitation is the scarcity of high-quality 3D scene data compared to the abundant image and video datasets. Additionally, existing 3D representations for generative models are either unstructured or memory-inefficient for large-scale scenes. In addition, previous methods typically generate monolithic 3D scenes where individual objects are not separated, limiting their applicability to interactive manipulations.

To address these fundamental limitations, we propose \shortname, a novel world generation framework that combines the best of both worlds, rather than treating 2D and 3D generation as separate paradigms. At the core of our approach is a semantically layered 3D mesh representation that enables structured 3D world generation with instance-level object modeling. Our method delivers three advanced features that distinguish it from existing approaches: (1) \textbf{360° immersive experiences} through panoramic world proxies that provide complete 360° scene coverage; (2) \textbf{mesh export capability} for seamless compatibility with existing computer graphics pipelines and industry-standard workflows; (3) \textbf{disentangled object representations} for object-level interaction within generated scenes.

To generate immersive and interactive 3D worlds with semantic layers, \shortname incorporates several key designs. First, we introduce \textit{panoramic world image generation} that serves as a unified world proxy for both text-to-world and image-to-world generation, leveraging the diversity of 2D generative models while providing immersive 360° environmental coverage. Second, we leverage \textit{agentic world layering}  for automating the decomposition of complex scenes into semantically meaningful layers, preparing for the subsequent layer-wise 3D reconstruction and disentangled object modeling. Third, we utilize \textit{layer-wise 3D world reconstruction}, which estimates aligned panoramic depth maps for generating a mesh-based 3D world across all extracted layers. The hierarchical world mesh representation contains explicitly separated objects while maintaining efficient memory usage and rendering performance. Finally, we present \textit{long-range world exploration} with a novel world-consistent video diffusion model and a world caching mechanism, facilitating user navigation through extensive unseen scene areas far beyond the original viewpoints.

Taken together, these innovations enable our framework to achieve state-of-the-art performance in generating immersive, explorable, and interactive 3D worlds across diverse artistic styles and vairous scene types. Extensive experiments demonstrate that our method achieves superior performance compared to existing approaches. Besides, \shortname supports versatile applications ranging from virtual reality and game development to physical simulation and object interaction. \shortname represents a significant step toward living out everyone's imagination on creating, exploring, and manipulating 3D worlds, bridging the gap between 2D content creation and immersive 3D experiences.

%% file: sections/method.tex
\section{Technical Details} 
\begin{figure}
\centering
\includegraphics[width=\textwidth]{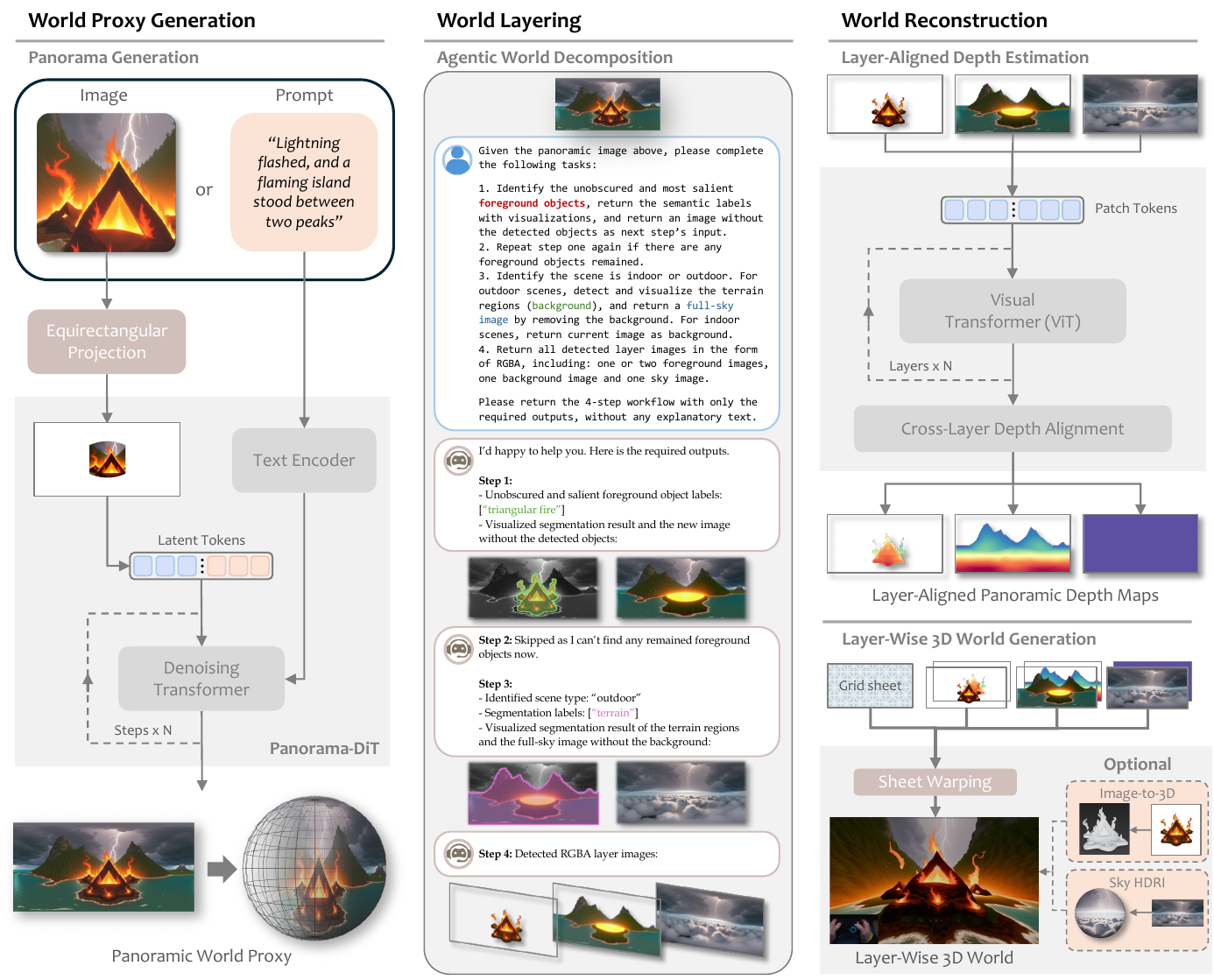}
\vspace{-3mm}
\caption{
An overview of \shortname architecture for 3D world generation. Given a conditioned scene image or textual description, \shortname generates layer-wise 3D worlds in mesh through a staged generative framework. We first leverage a diffusion model (Panorama-DiT) to generate a panoramic image, which serves as an initial world proxy for providing full 360° scene information. We then obtain semantically layered scene representations via world layering and reconstruction. To ensure layer-wise alignment of the reconstructed 3D world, we enhance the panoramic depth estimation model with a cross-layer depth alignment strategy. Also, users can obtain full 3D objects via image-to-3D generation or represent the sky as HDRI maps for downstream applications.
}
\label{fig_full_pipe}
\end{figure}

Our goal is to enable 3D world generation that supports both image and text inputs, adapting to diverse user needs. Compared to 3D objects, 3D worlds exhibit far greater diversity — encompassing indoor and outdoor environments, varying styles, and a wide range of scales, from a single room to an entire city. However, 3D scene data is scarce and hard to scale, making 3D world generation challenging. 
To address this, we propose combining 2D generative models with 3D generation by leveraging panoramas as a proxy representation for worlds. 
As illustrated in Fig.~\ref{fig_full_pipe}, \shortname is a staged generative framework, where we first utilize a diffusion model to generate a panorama as the world initialization, followed by world layering and reconstruction.  We detail the whole 3D world generation pipeline in the following subsections.

\input{sections/method/pano}

\input{sections/method/data}

\input{sections/method/agent}

\input{sections/method/recon}

\input{sections/method/video}

%% file: sections/method/pano.tex
\subsection{Generating Panoramas As World Proxy}
\label{sec:pano_gen}
Panoramas capture 360° visual information of a scene and can be formatted as equirectangular projection (ERP) images, making them an ideal proxy for 3D world generation. Thus, we generate a panorama as proxy for 3D world generation from text conditions or image conditions. 
 
\noindent \textbf{Text Conditions.} For text-to-panorama generation, the inputs are user-provided sentences. A critical challenge arises here: natural language inputs from users often differ significantly from the caption styles on which the models are trained.  To bridge this gap, we employ a LLM to translate the given text prompts if they are in Chinese, and then enhancing their details. This transformation ensures the prompts are well-aligned with the training data distribution of the generative model, thereby facilitating the generation of high-quality panoramas.

\noindent \textbf{Image Conditions.} Given a user-provided pinhole image, we aim generate coherent contents in the missing parts to obtain a complete 360° panorama while preserving the input image's content. To achieve this, we unproject the input image into panoramic space via equirectangular projection (ERP) with camera intrinsics estimated by a pretrianed 3D reconstruciton model, such as MOGE~\cite{wang2025moge} or UniK3D~\cite{piccinelli2025unik3d}.

\noindent \textbf{Panorama Generation.}
The architecture of our panorama generation model (Panorama-DiT) is based on the Diffusion Transformer (DiT) framework~\cite{peebles2023scalable}.  For text-to-panorama generation, only the enhanced text prompts are fed into the diffusion model as condition. For image-to-panorama generation, we first project the input image into panoramic space and then encode it into the latent space using a variational autoencoder (VAE). Next, we concatenate the condition image with the noisy latent for diffusion model. To enhance the generation quality and provide additional control, the image-to-panorama generation process is also conditioned on an auxiliary textual description produced by the scene-aware prompt generation method introduced in~\cref{sec:pano_data}.

Compared to general image generation, generating panoramic images faces unique challenges:  1) geometric distortion from spherical projection and 2) incontinuous boundary due to panoramic stitching. To address these issues, we introduce two key boundary artifact mitigation strategies: 1) elevation-aware augmentation. During training, we randomly shift ground-truth panoramas vertically (with probability $p$ and displacement ratio $r$) to enhance the robustness to viewpoint variations.
2) circular denoising~\cite{feng2023diffusion360}. During inference, we apply circular padding with progressive blending in denoising process to preserve structural and semantic continuity across panorama boundaries.

%% file: sections/method/data.tex
\subsection{Panoramic Data Curation Pipeline}
\label{sec:pano_data}

\begin{figure}[t]
    \centering
    \includegraphics[width=\linewidth]{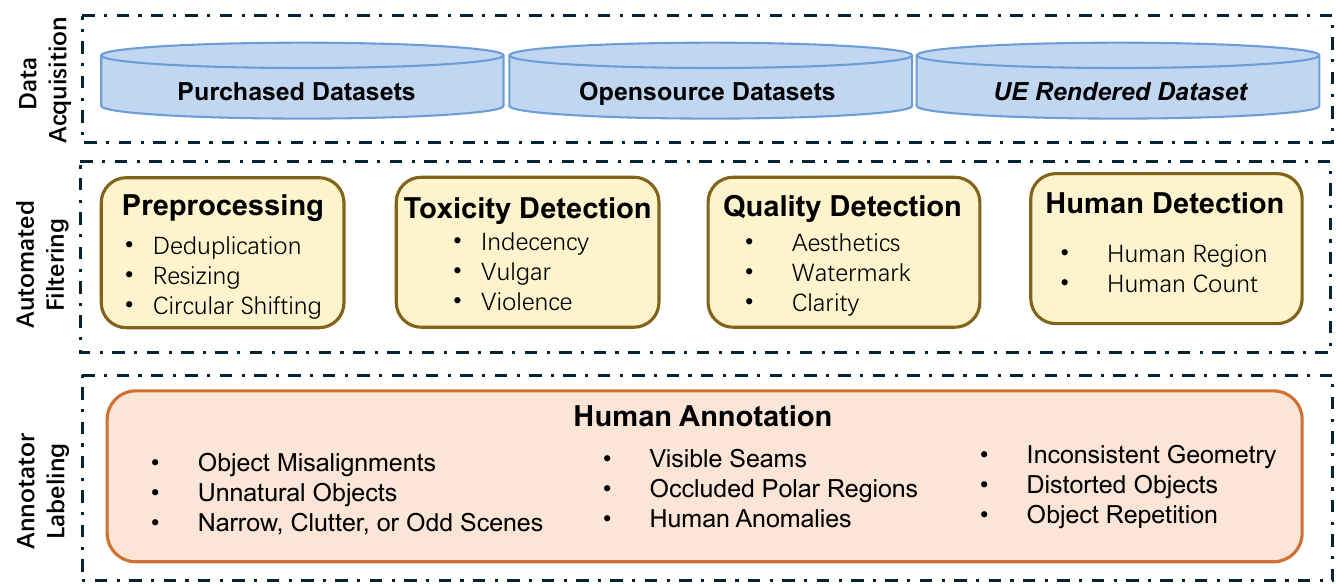}
    \vspace{-3mm}
    \caption{An overview of our panoramic data curation pipeline.}
    \label{fig:panorama_data_pipeline}
\end{figure}

\noindent \textbf{Data Curation.} The pipeline for our training data curation is illustrated in \figref{fig:panorama_data_pipeline}. Panoramic images are sourced from commercial acquisitions, open data downloads, and custom renders via Unreal Engine (UE).
Each panorama undergoes an automatic quality assessment framework, including watermark, aesthetic score, clarity, resolution, and distortion, \etc. Panoramas failing to meet predefined quality baselines will be discarded. We also invite expert annotators to manually inspect remaining samples, filtering examples with artifacts such as: 1) geometric artifacts (\eg, obvious distortion, visible boundary seam), 2) scene irregularities (\eg, narrow/unrepresentative spaces), and 3) content inconsistencies (\eg, abnormal object repetition, anomaly human bodies and objects).

\noindent \textbf{Training Caption.} 
Existing VLMs face challenges when generating captions for panoramic images, which contain richer visual details than general perspective images. It either generates overly simplified descriptions that fail to capture sufficient scene details or produces repetitive text with hallucinated elements. To mitigate these issues, we propose a three-stage captioning pipeline. We first leverage the re-captioning technique~\cite{li2024hunyuandit} to produce regularized descriptions with rich details for panoramas. We then utilize LLM to distill these descriptions into a collection of captions with varying lengths, spanning from high-level scene summaries to fine-grained object annotations. Finally, we invite professional annotators to verify the generated captions to eliminate image-text misalignment, ensuring semantic fidelity and minimizing hallucinations.

\noindent \textbf{Scene-Aware Prompt for Image Conditions.} As stataed in~\cref{sec:pano_gen}, for image-to-panorama generation, we utilize both image and text conditions. A straightforward approach to obtain the text prompt would be to use a vision-language model (VLM) to generate a caption for the input image. However, user-provided images often contain prominent objects (\eg, a statue), and conditioning the generation process on the input image's caption may lead to unwanted replication of these elements in the synthesized panorama (\eg, duplicate statues).

To address this, we introduce a scene-aware prompt generation strategy. We first instruct the VLM to identify salient objects in the input image and incorporate these as negative prompts to prevent the model from redundantly reproducing existing objects. We then instruct the VLM to envision a complete 360° scene that extends beyond the input image. Finally, we instruct the VLM to produce a refined and complete prompt that describe the scene hierarchically, from foreground to background and artistic styles to environmental atmosphere.
 
%% file: sections/method/agent.tex
\subsection{Agentic World Layering}
\label{subsec:agent}
While panoramas effectively serve as world proxies, they inherently lack information in occluded regions, supporting only viewpoint rotation rather than free exploration (\eg viewpoint translation). Inspired by human modeling practices, where artists typically model a 3D world scene as sky (sky box or dome), terrain mesh, and multiple object assets, we introduce a semantically layered 3D world representation that decomposes a scene into a sky layer (for outdoor scenes), a background layer, and multiple object layers. To automate the layering process, we develop an agentic world decomposition method consisting of instance recognition, layer decomposition, and layer completion.

\noindent \textbf{Instance Recognition.}
In the context of a playable 3D world, interaction with specific scene objects is essential, while background elements typically remain static. To enable an interactive 3D scene, we must model each interactive object individually in 3D. Our approach begins with identifying which scene objects require such modeling, which demands semantic understanding and spatial relationship reasoning.   Given the diversity of our generated scenes—spanning indoor/outdoor environments, natural landscapes, and game-style settings, we leverage a VLM to harness its rich world knowledge for semantic object recognition.
Following instance recognition, we further categorize objects into multiple sub-layers based on their semantic and spatial relationships. For instance, in an urban scene, we target segregating nearby vehicles from distant buildings into separate layers.
 
\noindent \textbf{Layer Decomposition.}  Once obtaining semantic labels, the next step is to determine the precise positions of these recognized objects. However, conventional visual grounding models cannot be directly applied to panoramas due to their inherent spatial discontinuities, where objects may be fragmented across the left and right boundaries of an equirectangular projection panorama.

To address this, we preprocess the panoramic image using circular padding before inputting it to an object detector (\eg, Grounding DINO~\cite{liu2024grounding}). This transformation ensures that objects spanning the panorama's boundaries are treated as contiguous entities. Following detection, we remap the bounding box coordinates from the padded space back to the original panorama. Subsequently, we pass the detected bounding boxes to a segmentation model (\eg, ZIM~\cite{kim2024zim}) to generate pixel-wise masks. To handle overlapping or fragmented detections, we apply non-maximum suppression (NMS) based on object area size. This ensures that part-level objects separated by the panorama's boundaries are merged, enabling more accurate 3D modeling of interactive scene elements.

\noindent \textbf{Layer Completion.} 
Following object segmentation, we decompose the panorama into background (\eg, terrain) and sky layers through an autoregressive process. This involves iteratively removing recognized objects and completing each layer by inpainting occluded regions.
 
To train a layer completion model for panoramic scenes, we curate a panoramic object removal dataset consisting of triples of a object mask, the original panorama containing the object, and a target panorama with the object removed. Using this dataset, we fine-tune our Panorama-DiT model to learn faithful content completion of occluded regions. Similarly, we also finetune a specific completion model for sky layer on a dataset of sky HDRIs.

%% file: sections/method/recon.tex
\subsection{Layer-Wise World Reconstruction}
\label{subsec:recon}
Given the hierarchical world layers, we reconstruct the 3D world in a layer-wise manner. As shown in~\figref{fig_full_pipe} (right), the reconstruction process includes two stages: (1) layer-aligned depth estimation and (2) layer-wise 3D world generation.

\noindent \textbf{Layer-Aligned Depth Estimation.} Given the panoramic world proxies, we predict the depth of each layer and then conduct cross-layer depth alignment. Specifically, we first obtain a base depth map by applying the depth estimation model~\cite{wang2025moge,piccinelli2025unik3d} to the original panorama. The depth of objects in the first foreground layer can be extracted from the base panoramic depth map. 

For subsequent layers (\eg, the subsequent foreground layers and the background layer after foreground removal), we predict their depths separately and align them with the base panoramic depth map using depth matching techniques that minimizing the distance of overlapped regions across different layers. This ensures consistent depth relationships across different layers to maintain the geometric coherence of the reconstructed 3D scene.
For the sky layer, we set its depth to a constant value that slightly lager than the maximum depth value observed across all existing layers, ensuring the sky appears at the farthest distance.

\noindent \textbf{Layer-Wise 3D World Generation.} Given the layered  images with aligned depth maps, we reconstruct the world via sheet warping with a grid mesh representation in WorldSheet~\cite{hu2021worldsheet}. The reconstruction process follows a hierarchical approach.

\noindent \textit{Foreground Object Reconstruction.} For each foreground layer, we offer two reconstruction strategies: (1) \textit{Direct projection}, where we convert the foreground objects directly to 3D meshes via sheet warping based on their depths and semantic masks. To ensure the quality of meshes warped from a panoramic image with masks, we also introduce special handling for polar region smoothing and mesh boundary anti-aliasing; (2) \textit{3D generation}, where we generate complete 3D objects in the foreground layers and then place them into the 3D world. To obtain the foreground 3D objects, we extract individual object instances from the foreground layers based on their instance masks, and leverage image-to-3D generation models (\eg, Hunyuan3D~\cite{yang2024hunyuan3d, hunyuan3d22025tencent, lai2025hunyuan3d25highfidelity3d}) to create high-quality 3D object assets. We also propose an automatic object placement algorithm to place generated objects into 3D scenes considering spatial layout.

\noindent \textit{Background Layer Reconstruction.} for background layer, we first apply adaptive depth compression to handle depth outliers and ensure proper depth distribution. We then convert the background panoramic image into a 3D mesh via sheet warping using the processed background depth map.

\noindent \textit{Sky Layer Reconstruction.} The sky layer is reconstructed using the sky image with uniform depth values set to be slightly larger than the maximum scene depth. In addition to traditional mesh representation obtained via sheet warping, we also support HDRI environment map representation for more realistic sky rendering in VR applications.

We also support 3D gaussian splatting as an alternative to the mesh representation by optimizing a layered 3DGS representation based on the depth.
To handle cross-boundary consistency in equirectangular projections, we apply circular padding during the reconstruction process, ensuring seamless transitions at the panorama boundaries. The final layered 3D world maintains proper occlusion relationships and depth ordering, enabling realistic VR experiences with proper parallax effects.

%% file: sections/method/video.tex
\subsection{Long-Range World Extension} \label{subsec:video} While layer-wise world reconstruction enables world exploration, challenges remain with occluded views and limited exploration range. To address these limitations, we introduce Voyager~\cite{huang2025voyager}, a video-based view completion model that enables consistent   world extrapolation. Voyager combines world-consistent video diffusion with long-range exploration mechanisms to synthesize spatially coherent RGB-D videos from an initial world view and user-specified camera trajectories.

\noindent\textbf{World-Consistent Video Diffusion.} Voyager employs an expandable world caching mechanism to maintain spatial consistency and prevent visual hallucination. The system constructs an initial 3D point cloud cache with the generated 3D scene, then projects this cache into target camera views to provide partial guidance for the diffusion model. The generated frames continuously update and expand the world cache, creating a closed-loop system that supports arbitrary camera trajectories while preserving geometric coherence.

\noindent\textbf{Long-Range World Exploration.} To overcome the limitations of generating long videos in a single pass, we propose a world caching scheme combined with smooth video sampling for auto-regressive scene extension. The world cache accumulates point clouds from all generated frames, with a point culling method that removes redundant points to optimize memory usage. Using cached point clouds as spatial proxies, we develop a smooth sampling strategy
that auto-regressively extends video sequences while ensuring seamless transitions between clips.

\subsection{System Efficiency Optimization}

To ensure practical deployment and real-time performance, HunyuanWorld 1.0 incorporates comprehensive system optimizations across both mesh storage and model inference components.

\noindent\textbf{Mesh Storage Optimization.} Meshes for a 3D scene are large for loading and storing. We thus employ dual compression strategies for both offline usage and online deployment scenarios to achieve efficient storage and fast loading while maintaining visual quality.

\textit{Mesh Decimation with Advanced Parameterization.} For offline mesh usage, we employ a multi-stage pipeline consisting of mesh decimation, texture baking, and UV parameterization. We evaluate an XAtlas-based solution~\cite{xatlas} for UV parameterization, which keeps good UV quality while eliminating rendering seams compared with naive parameterization methods. The compression pipeline achieves 80\% size reduction, making it suitable for high-quality offline content preparation despite extended processing times.

\noindent\textit{Web-Optimized Draco Compression.} For online web deployment scenarios, we adopt the Draco~\cite{draco}, which delivers exceptional compression efficiency while preserving visual fidelity. This approach demonstrates superior size reduction (90\%) capabilities and maintains rendering quality comparable to uncompressed meshes. The format provides native WebAssembly support, ensuring seamless integration with web-based graphics pipelines and broad browser compatibility. 

\noindent\textbf{Model Inference Acceleration.} Our inference optimization employs a comprehensive TensorRT-based acceleration framework with intelligent caching and multi-GPU parallelization. The system converts diffusion transformer models into optimized TensorRT engines, supporting both cached and uncached inference modes with shared memory allocation to minimize GPU overhead. We implement a selective caching strategy that applies cached inference for non-critical denoising steps while using full computation for key steps that significantly impact generation quality. For classifier-free guidance scenarios, the system leverages multi-GPU parallel processing through threaded execution, simultaneously computing positive and negative prompt conditions on separate devices with synchronized result aggregation. 
This integrated optimization approach enables fast 3D world generation while maintaining high visual quality across diverse deployment environments.

%% file: sections/experiments.tex
\begin{figure}[t]
\centering 
\includegraphics[width=\textwidth]{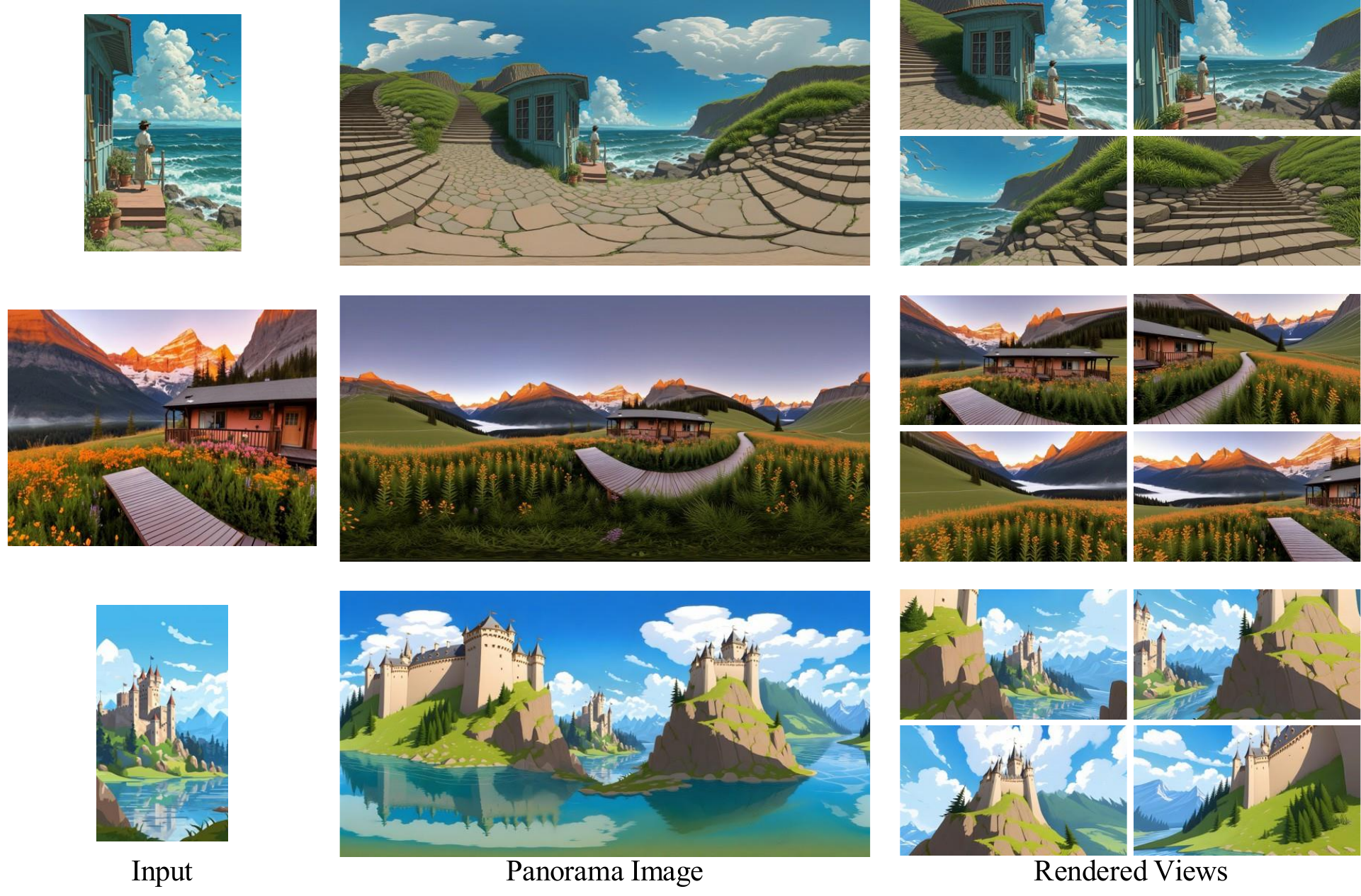}
\vspace{-4mm}
\caption{
Visual results of image-to-panorama generation by \shortname.
}
\label{fig:pano_good_case_i2p}
\end{figure}

 \input{tables/panorama/i2p}
 
\begin{figure}[t]
\centering 
\includegraphics[width=\textwidth]{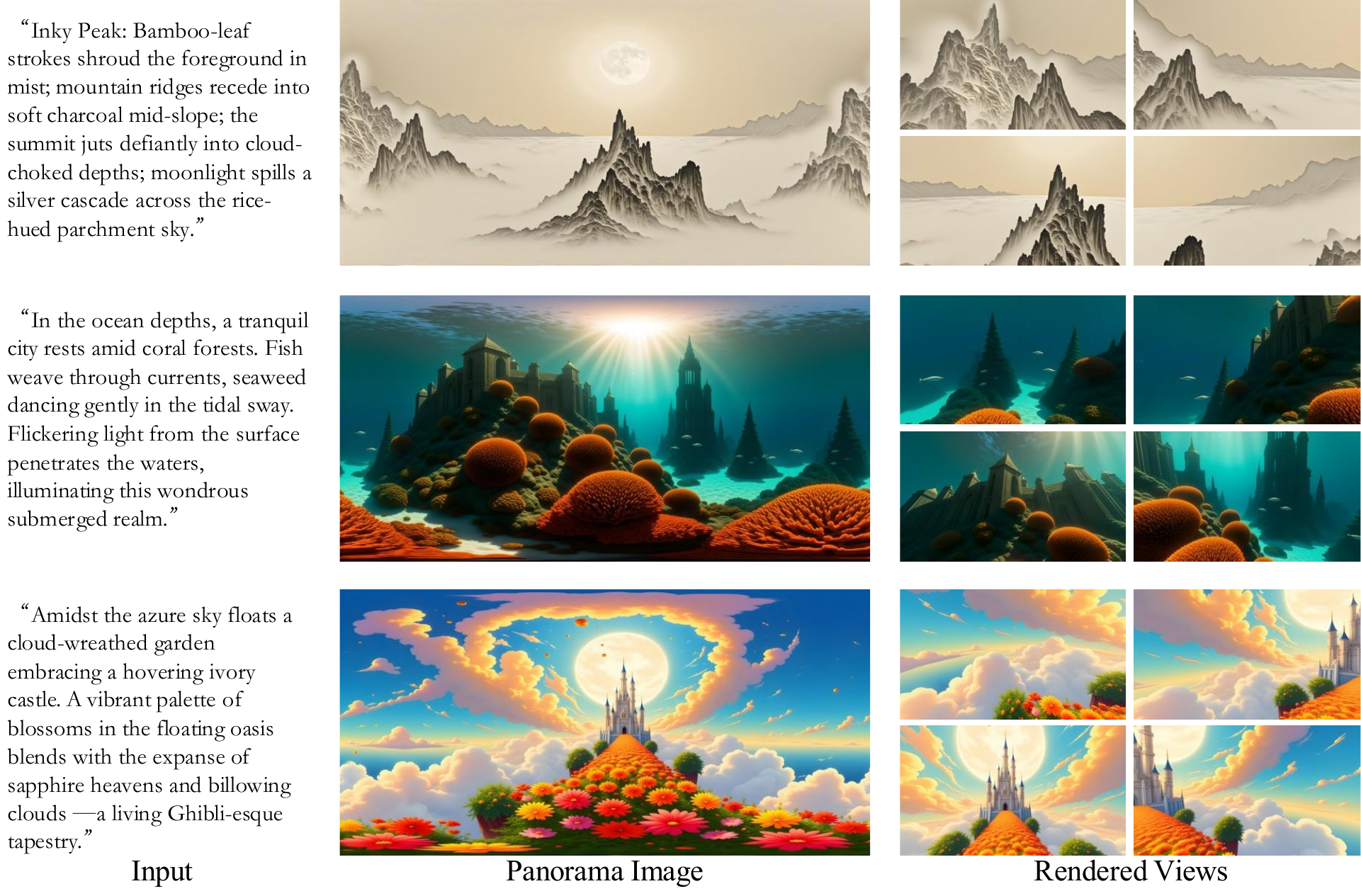}
\vspace{-4mm}
\caption{
Visual results of text-to-panorama generation by \shortname.
}
\label{fig:pano_good_case_t2p}
\end{figure}

\section{Model Evaluation}
\begin{figure}[t]
\centering 
\includegraphics[width=\textwidth]{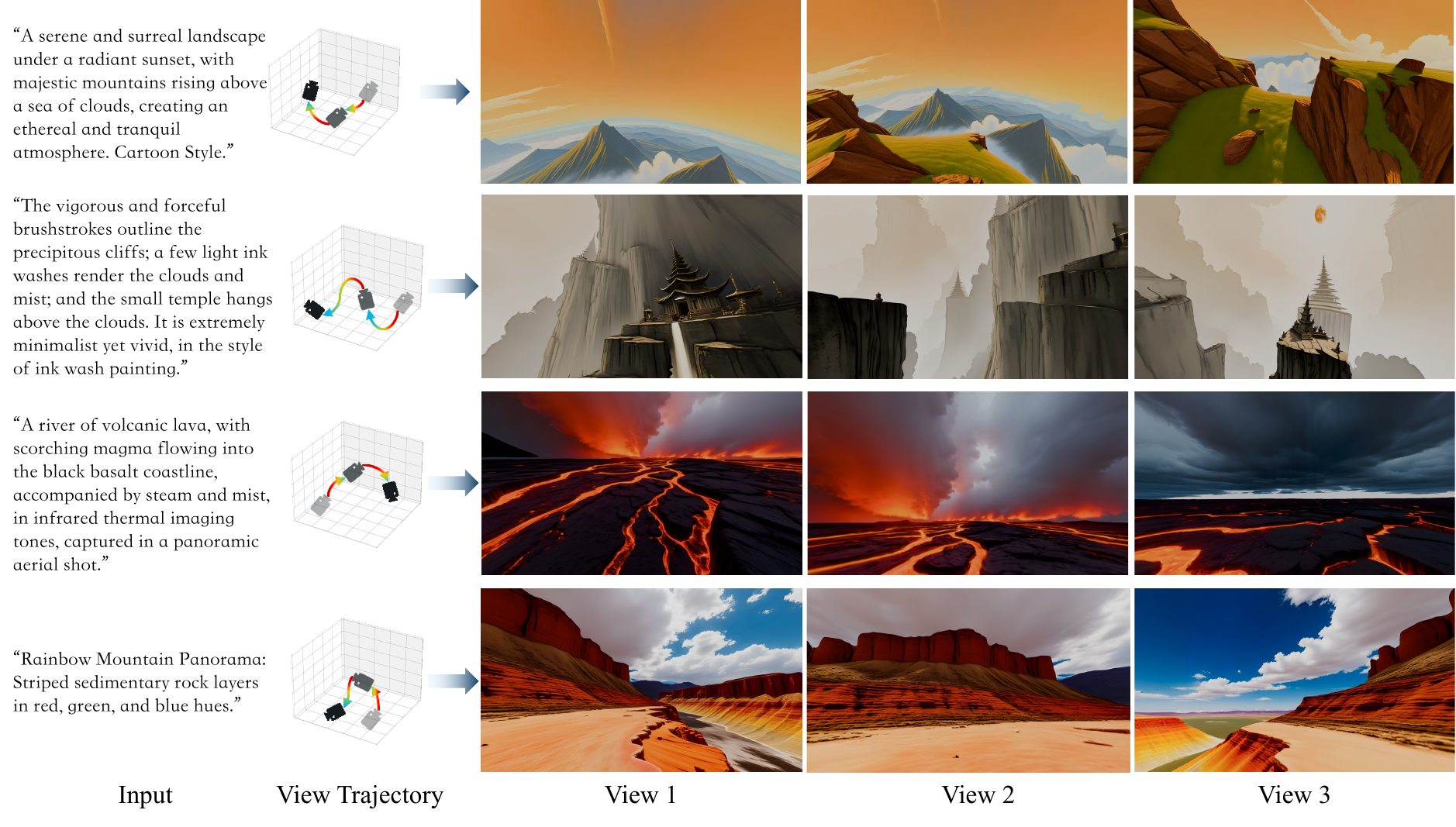} 
\vspace{-4mm}
\caption{
Visual results of text-to-world generation by \shortname.
}
\label{fig:t2w_good_case}
\end{figure}

\begin{figure}[t]
\centering 
\includegraphics[width=\textwidth]{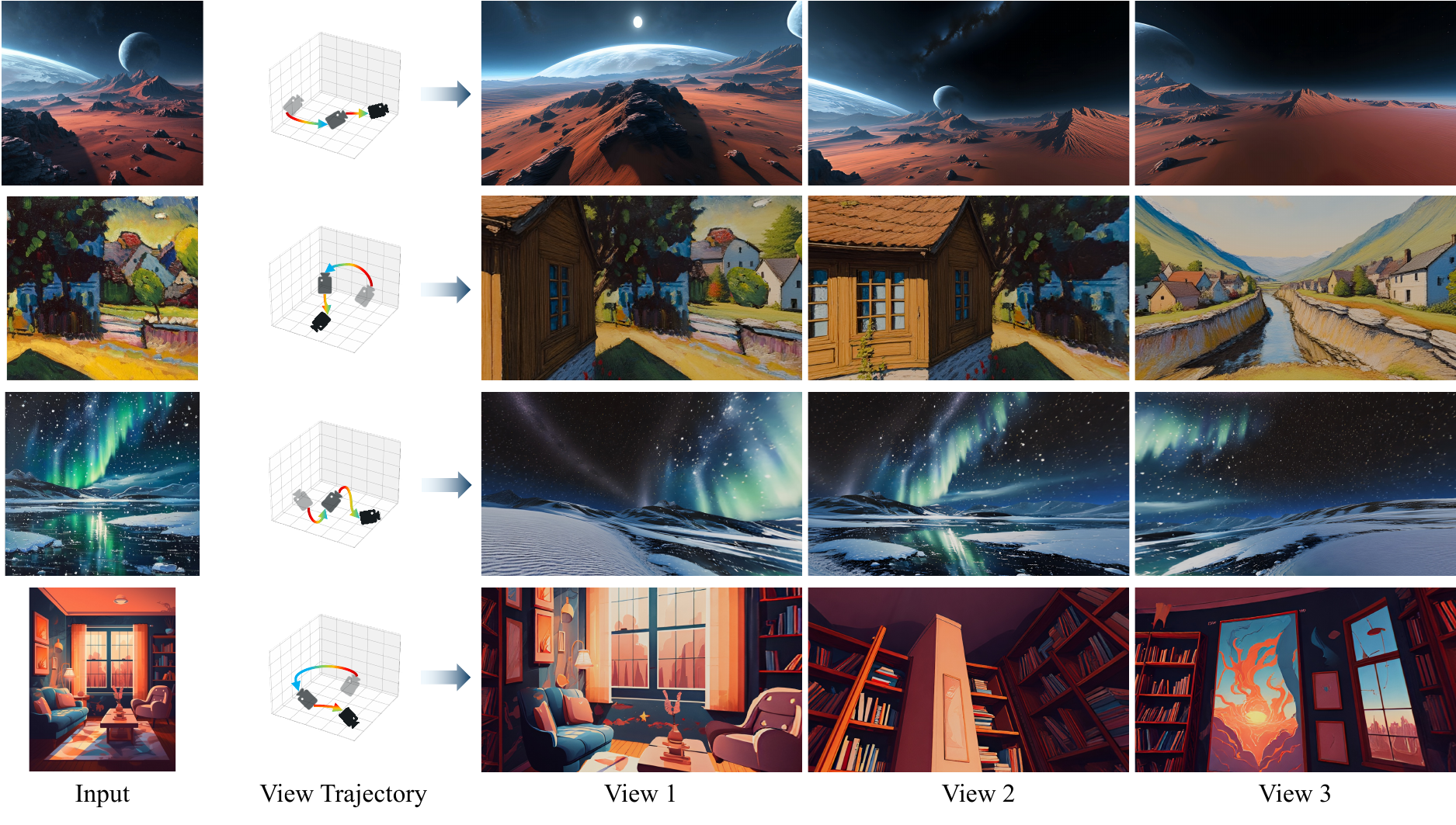} 
\vspace{-4mm}
\caption{
Visual results of image-to-world generation by \shortname.
}
\label{fig:i2w_good_case}
\end{figure}

We show the generated panoramic images and explorable 3D worlds from image and text input in~\figref{fig:pano_good_case_i2p}, \ref{fig:pano_good_case_t2p}, 
\ref{fig:t2w_good_case}, \ref{fig:i2w_good_case}. We can see that \shortname generates high-quality panoramic images that precisely follow the conditions, while the generated 3D worlds maintain spatial coherence and enable immersive exploration across diverse camera trajectories, scene types, and artistic styles.
For the rest of this section, we first conduct experiments to compare our results with those generated by the state-of-the-art methods on both panorama generation and 3D world generation. We then introduce a series of practical applications to demonstrate the versatility of \shortname on virtual reality, physical simulation, game development, and interactive object manipulation scenarios.

\subsection{Evaluation Protocol}

\textbf{Benchmark.}
For image-conditioned generation, we evaluate on both real and AI-generated images, curating from World Labs~\cite{worldlabs}, Tanks and Temples~\cite{knapitsch2017tanks}, and images collected from real users.
For text-conditioned generation, we curate a prompt benchmark by crowd-sourcing, covering different scene types, styles, and lengths.

\textbf{Metrics.}
Due to the absence of ground-truth data for comparisons, we follow the setting of  \cite{schwarz2025recipe} to quantitatively assess the performance of our method from two key aspects: the alignment between input and output, alongside the visual quality. 
To evaluate the input-output alignment, we utilize the CLIP score~\cite{hessel2021clipscore} to measure the similarity of the generated worlds and the given prompts (CLIP-T) or images (CLIP-I).
To evaluate the visual quality, we employ a few non-reference image quality assessment metrics, including BRISQUE~\cite{mittal2012no}, NIQE~\cite{mittal2012making}, and Q-Align~\cite{wu2023q}.

\input{sections/experiments/panorama}

\input{sections/experiments/3d-worlds}

\input{sections/experiments/applications}

%% file: tables/panorama/i2p.tex
\begin{table}[t]
\centering
\small
\begin{tabular}{rcccccccc}
\hline
            & BRISQUE ($\downarrow$) & NIQE ($\downarrow$) & Q-Align ($\uparrow$) & CLIP-I ($\uparrow$)    \\ \hline
Diffusion360~\cite{feng2023diffusion360}      & 71.4    & 7.8     & 1.9    & 73.9         \\
MVDiffusion~\cite{Tang2023mvdiffusion}     & \underline{47.7}    & \underline{7.0}     & \underline{2.7}    & \underline{80.8}         \\  
\shortname (Ours)  & \textbf{45.2}    & \textbf{5.8}     & \textbf{4.3}    & \textbf{85.1}           \\        \hline
\end{tabular}
\vspace{3mm}
\caption{
Quantitative comparisons for image-to-panorama generation.
}
\label{tab:i2p}
\end{table}

%% file: sections/experiments/panorama.tex
\input{tables/panorama/t2p}

\subsection{Panorama Generation}
 
We compare and evaluate both image-to-panorama generation and text-to-panorama generation.

\textbf{Image-to-Panorama Comparisons.}
\textit{Settings.}
We compare \shortname with two state-of-the-art image-based panorama generation methods, Diffusion360~\cite{feng2023diffusion360} and MVDiffusion~\cite{Tang2023mvdiffusion}.
We measure the visual quality of the generated panoramic images and the similarity between the CLIP image embeddings of novel images rendered from the generated panorama and the input image. Specifically, we render six views with $90^\circ$ field of view (FOV), strategically positioned to provide complete $360^\circ$ coverage. Each view is rendered at a resolution of $960\times960$.

\begin{figure}
\centering
\includegraphics[width=0.9\textwidth]{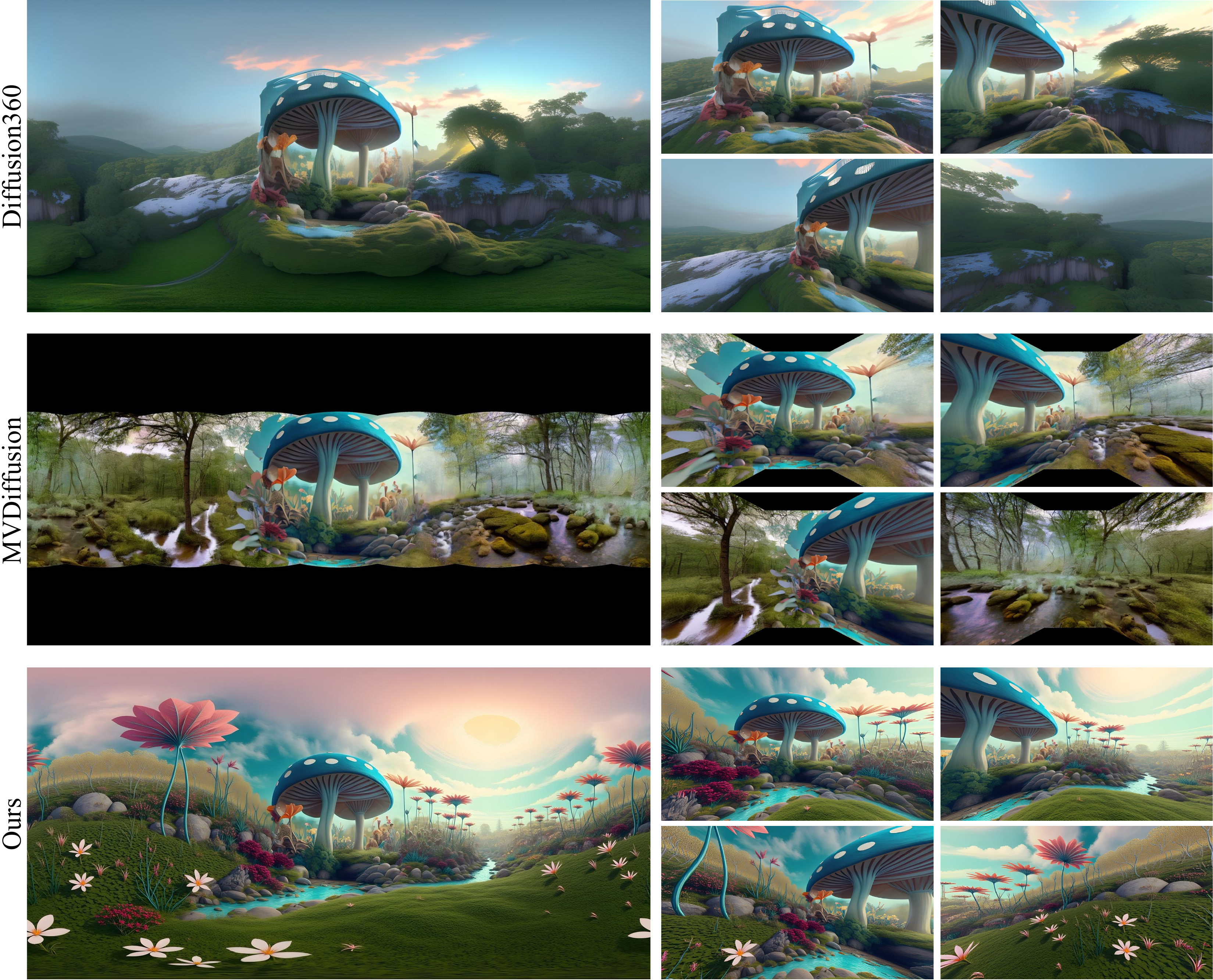}
\caption{
Qualitative comparisons for image-to-panorama generation (World Labs). Left: panoramic images generated from the same input image. Right: Four perspectively rendered views.}
\label{fig:i2p_cmp_1}
\end{figure}

\begin{figure}
\centering
\includegraphics[width=0.9\textwidth]{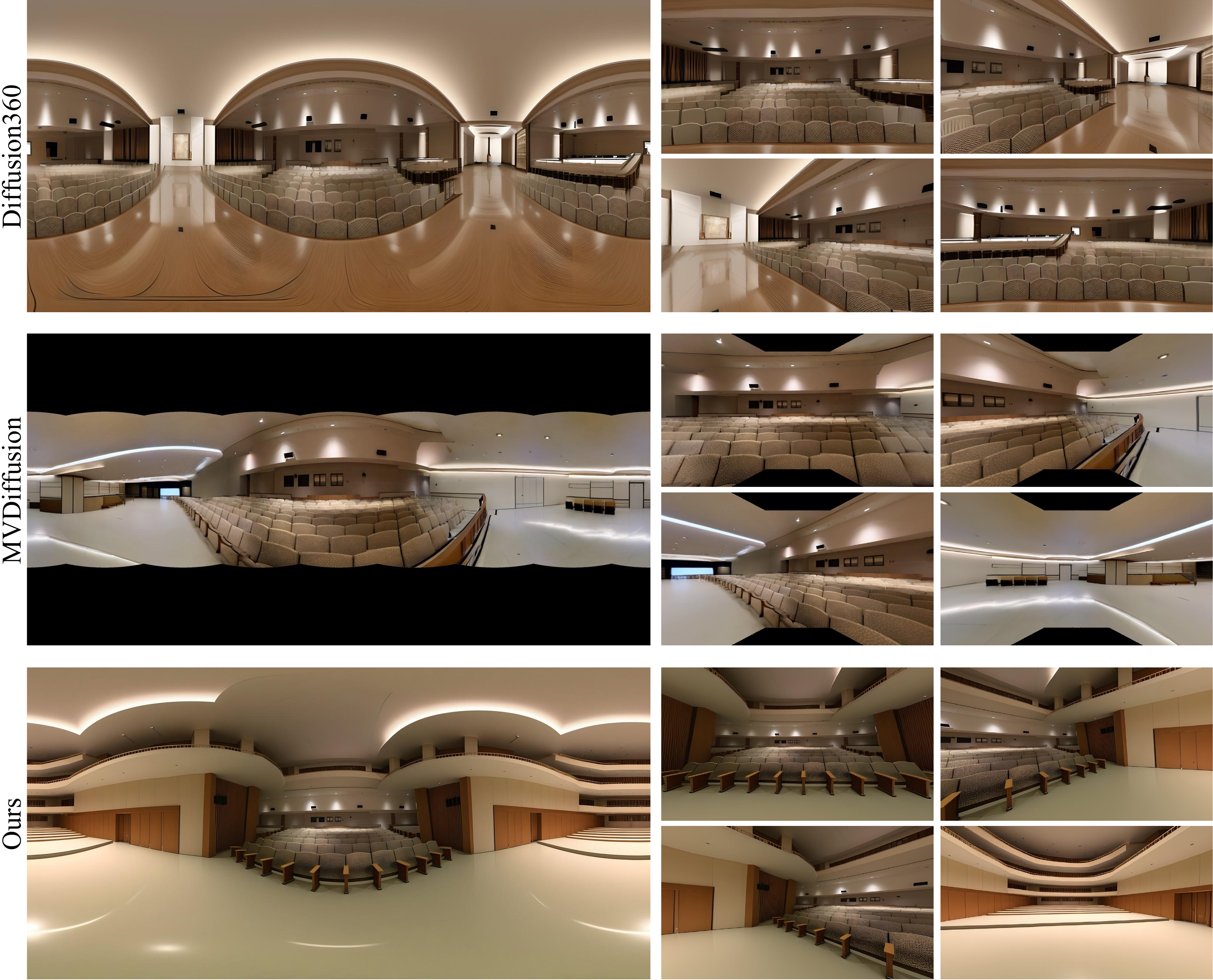}

\caption{
Qualitative comparisons for image-to-panorama generation (Tanks and Temples). Left: panoramic images generated from the same input image. Right: Four perspectively rendered views.
}
\label{fig:i2p_cmp_2}
\end{figure}

\textit{Results.}
The quantitative results presented in~\cref{tab:i2p} demonstrate that \shortname consistently outperforms both baseline methods across all evaluation metrics. These findings highlight the effectiveness of our method in producing high-fidelity panoramic images while preserving strong semantic correspondence with the input.
Qualitative comparisons illustrated in~\cref{fig:i2p_cmp_1} and~\cref{fig:i2p_cmp_2} corroborate these quantitative findings. In contrast to baseline approaches, which frequently exhibit discontinuous artifacts and geometric distortions, our method generates panoramic scenes with enhanced visual coherence and aesthetic quality.

\begin{figure}
\centering 
\includegraphics[width=\textwidth]{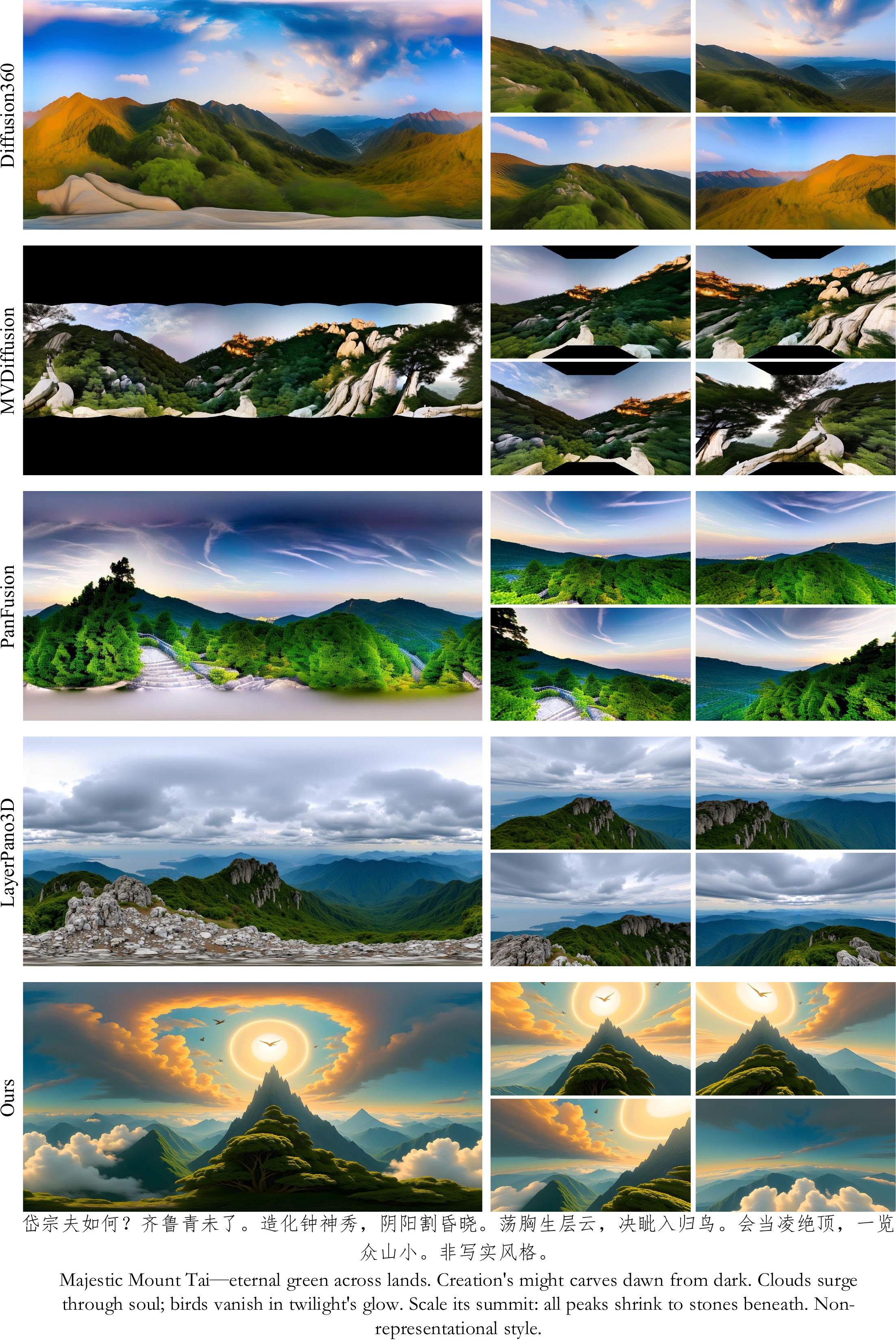}
\vspace{-3mm}
\caption{
Qualitative comparisons for text-to-panorama generation (case 1).
Left: panoramic images generated from the text at the bottom. Right: Four perspectively rendered views.
}
\label{fig:t2p_cmp_1}
\end{figure}

\begin{figure}
\centering
\includegraphics[width=\textwidth]{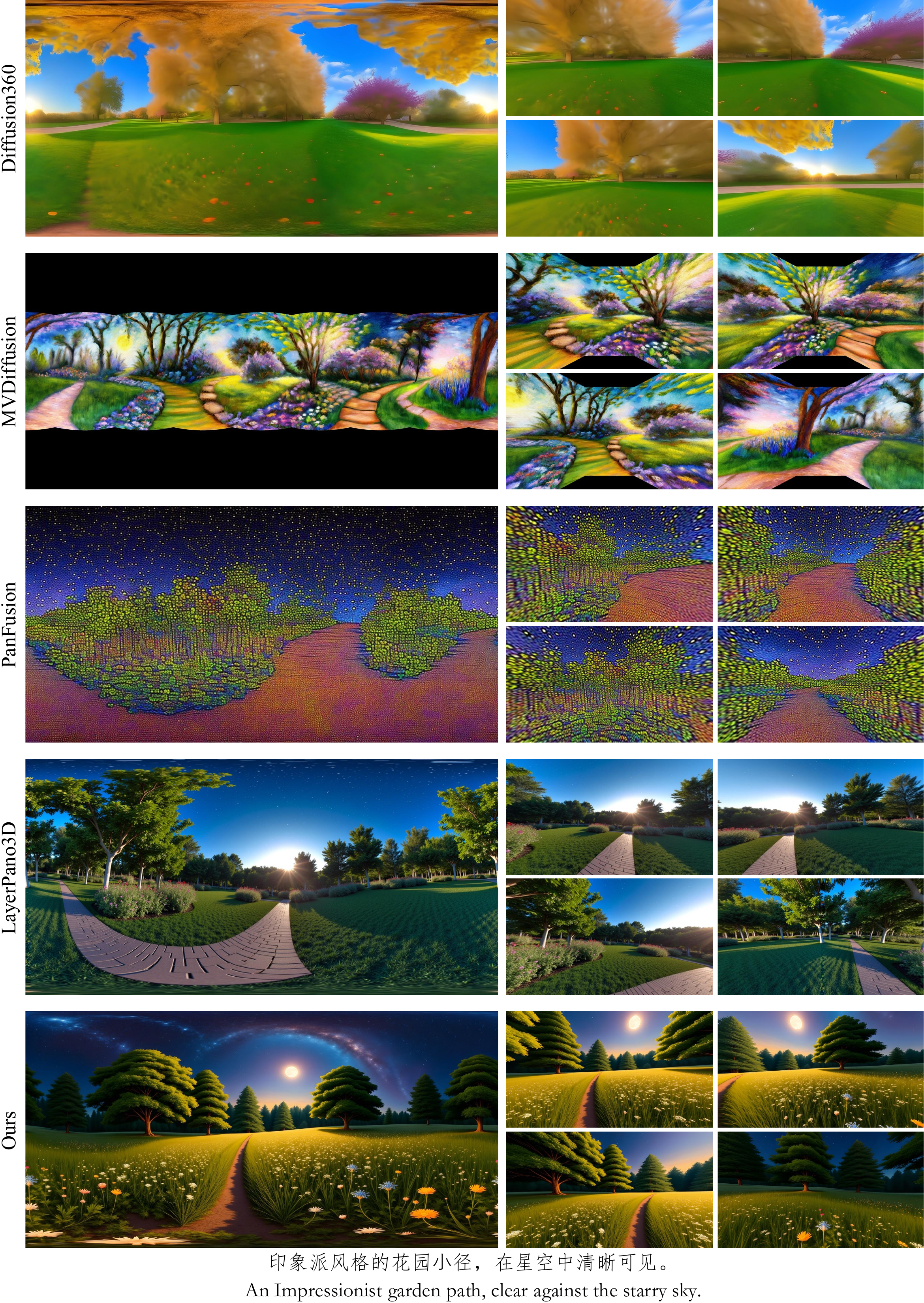}
\vspace{-3mm}
\caption{
Qualitative comparisons for text-to-panorama generation (case 2).
Left: panoramic images generated from the text at the bottom. Right: Four perspectively rendered views.
}
\label{fig:t2p_cmp_2}
\end{figure} 
\textbf{Text-to-Panorama Comparisons.}
\textit{Settings.} We compare \shortname against four state-of-the-art text-conditioned panorama generation methods: Diffusion360~\cite{feng2023diffusion360}, MVDiffusion~\cite{Tang2023mvdiffusion}, PanFusion~\cite{zhang2024taming}, and LayerPano3D~\cite{yang2024layerpano3d}.
We maintain a consistent rendering strategy with the image-to-panorama experimental settings with six rendered perspective views at $90^\circ$ FOV and a resolution of $960\times960$ for each generated panoramic image. We utilize CLIP-T scores to quantify the semantic alignment between the generated panoramic image and the corresponding textual input.

\textit{Results.} 
The quantitative results in~\cref{tab:t2p} demonstrate that \shortname achieves superior performance across all evaluation metrics compared to the baseline methods. Qualitative results in~\cref{fig:t2p_cmp_1} and~\cref{fig:t2p_cmp_2} reveal that our approach exhibits exceptional fidelity to textual descriptions while maintaining high visual quality standards. Furthermore, \shortname excels at generating panoramic scenes across diverse artistic styles and thematic contexts.

%% file: tables/panorama/t2p.tex
\begin{table}[t]
\centering
\small
\begin{tabular}{rcccccccc}
\hline
            & BRISQUE ($\downarrow$) & NIQE ($\downarrow$) & Q-Align ($\uparrow$) & CLIP-T ($\uparrow$)    \\ \hline
Diffusion360~\cite{feng2023diffusion360}      & 69.5    & 7.5     & 1.8    & 20.9         \\ 
MVDiffusion~\cite{Tang2023mvdiffusion}     & \underline{47.9}    & 7.1     & 2.4    & \underline{21.5}         \\ 
PanFusion~\cite{zhang2024taming}     & 56.6    & 7.6     & 2.2    & 21.0         \\ 
LayerPano3D~\cite{yang2024layerpano3d}       & 49.6    & \underline{6.5}     & \underline{3.7}    & \underline{21.5}         \\
\shortname (Ours)  & \textbf{40.8}    & \textbf{5.8}     & \textbf{4.4}    & \textbf{24.3}           \\        \hline
\end{tabular}
\vspace{3mm}
\caption{
Quantitative comparisons for text-to-panorama generation.
}
\label{tab:t2p}
\end{table}

%% file: sections/experiments/3d-worlds.tex
\input{tables/3d-world/i2w}
\input{tables/3d-world/t2w}

\begin{figure}
\centering 
\includegraphics[width=\textwidth]{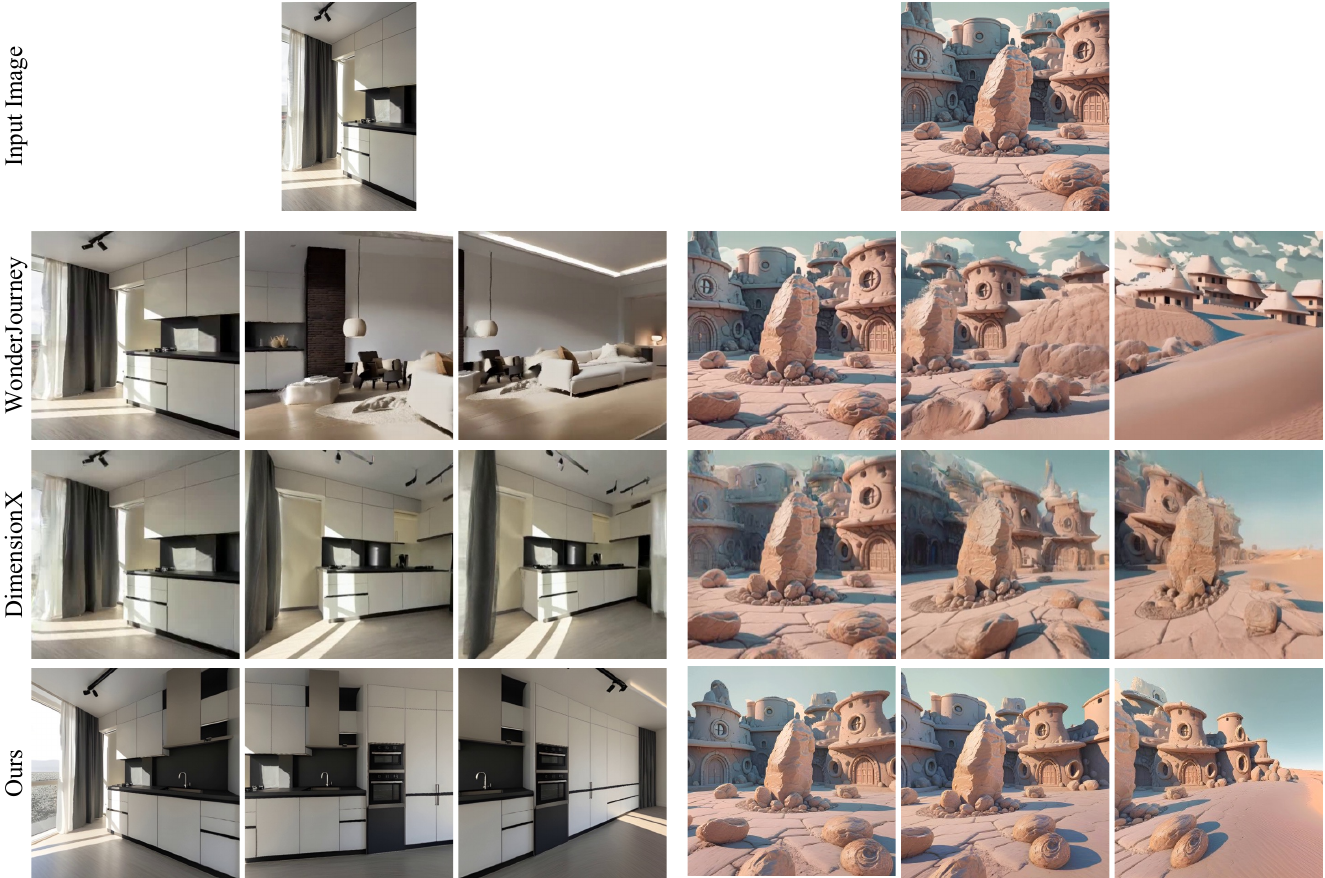} 
\caption{
Qualitative comparisons for image-to-world generation. For each case, we render three perspective views from the generated 3D scenes. 
}
\label{fig:i2w_cmp}
\end{figure}

\begin{figure}
\centering 
\includegraphics[width=\textwidth]{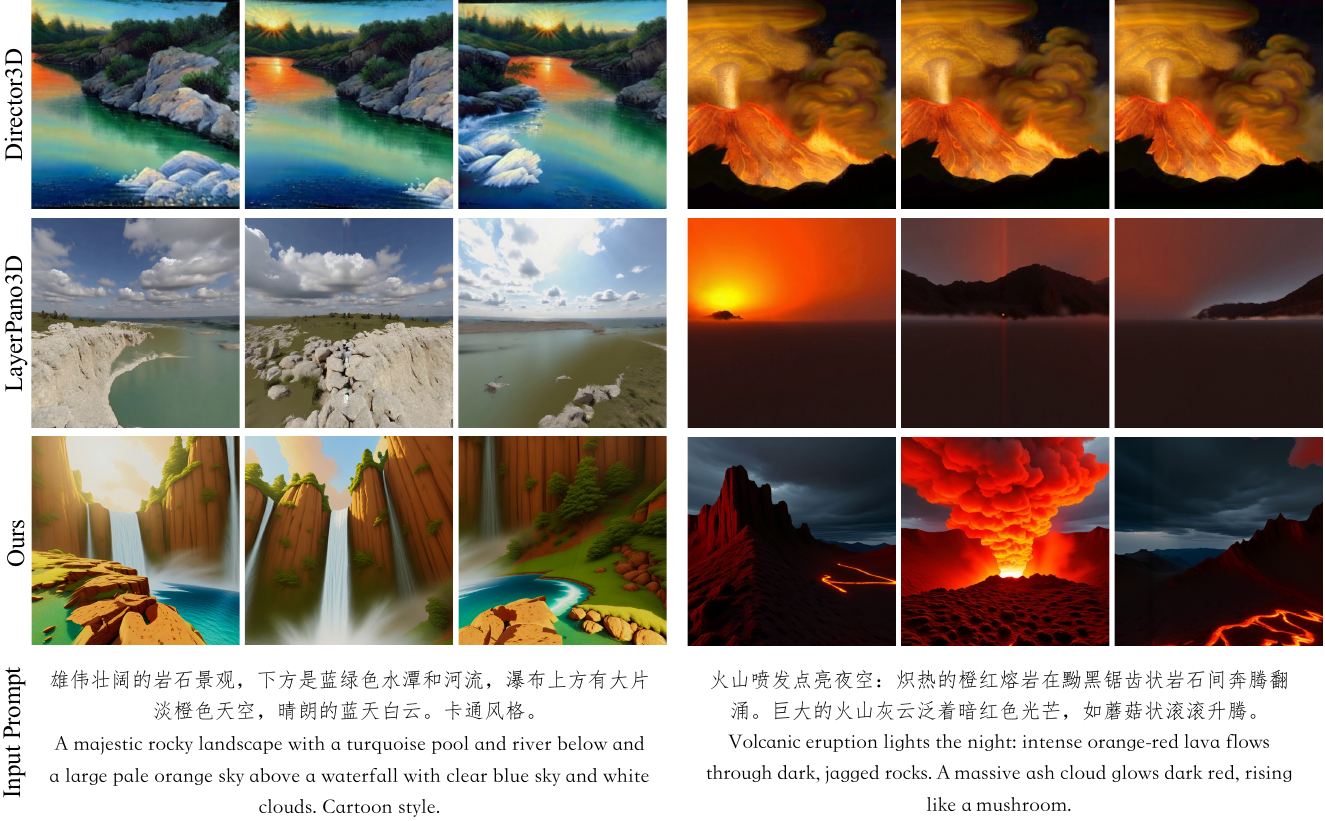} 
\caption{
Qualitative comparisons for text-to-world generation. For each case, we render three perspective views from the generated 3D scenes. 
}
\label{fig:t2w_cmp}
\end{figure}

\subsection{3D World Generation}
\label{exp:tex}

We compare and evaluate both image-to-world generation and text-to-world generation capabilities of \shortname.

\textbf{Image-to-World Comparisons.}
\textit{Settings.}
We compare \shortname with two state-of-the-art image-based 3D world generation methods: WonderJourney~\cite{yu2024wonderjourney} and DimensionX~\cite{sun2024dimensionx}.
We measure both the visual quality of the generated 3D worlds and the alignment between the rendered novel views and the input images. For \shortname, we generate and reconstruct the complete 3D scenes for rendering seven views with $90^\circ$ FOV at  azimuth angles of $\{0^\circ,15^\circ,30^\circ,45^\circ,60^\circ,75^\circ,90^\circ\}$. For DimensionX, we utilize its open-source \textit{orbit} LoRA for video generation and 3D reconstruction, following its predefined $90^\circ$ orbital camera trajectory. For WonderJourney, we employ a camera trajectory that also rotates right $90^\circ$ for evaluation.  All novel views are rendered at a resolution of $960\times960$.

\textit{Results.}
The quantitative results presented in~\cref{tab:i2w} demonstrate that \shortname consistently outperforms both baseline methods on the visual quality of rendered novel views and semantic alignment with the input image.
Qualitative comparisons illustrated in~\cref{fig:i2w_cmp} show our method generates 3D worlds with superior visual quality and geometric consistency compared to the baseline approaches while maintaining enhanced alignment with the input images.

\textbf{Text-to-World Comparisons.}
\textit{Settings.}
We compare \shortname against two state-of-the-art text-conditioned 3D world generation methods: LayerPano3D~\cite{yang2024layerpano3d} and Director3D~\cite{li2024director3d}.
For a fair evaluation, we utilize a consistent evaluation protocol for \shortname and LayerPano3D by rendering six views with $90^\circ$ FOV at azimuth angles of \( \left\{0^\circ,60^\circ,120^\circ,180^\circ,240^\circ,300^\circ\right\} \) from the reconstructed 3D scenes. For Director3D, we utilize its model-predicted camera trajectories for novel view rendering, as its performance heavily depends on its self-predicted camera trajectories. We utilize CLIP-T scores to quantify semantic alignment between generated 3D content and its textual descriptions. All novel views are rendered at a resolution of $960\times960$.

\textit{Results.}
We present the quantitative results in~\cref{tab:t2w}, with qualitative comparisons shown in~\cref{fig:t2w_cmp}. \shortname consistently outperforms both baseline methods across all evaluation metrics. The visual results demonstrate that our approach generates 3D worlds with high visual fidelity and strong semantic alignment with the input text descriptions. Notably, Director3D exhibits limitations in generating long-range camera trajectories for many test cases, which restricts its ability to generalize across diverse input conditions.

%% file: tables/3d-world/i2w.tex
\begin{table}[t]
\centering
\small
\begin{tabular}{rcccccccc}
\hline
            & BRISQUE ($\downarrow$) & NIQE ($\downarrow$) & Q-Align ($\uparrow$) & CLIP-I ($\uparrow$)\\ \hline
WonderJourney~\cite{yu2024wonderjourney}      & 51.8 & 7.3 & 3.2 & 81.5 \\ 
DimensionX~\cite{sun2024dimensionx}       & \underline{45.2} & \underline{6.3} & \underline{3.5} & \underline{83.3}         \\ 
\shortname (Ours)  & \textbf{36.2}    & \textbf{4.6}     & \textbf{3.9}   & \textbf{84.5}        \\        \hline
\end{tabular}
\vspace{3mm}
\caption{
Quantitative comparisons for image-to-world generation.
}
\label{tab:i2w}
\end{table}

%% file: tables/3d-world/t2w.tex
\begin{table}[t]
\centering
\small
\begin{tabular}{rcccccccc}
\hline
            & BRISQUE ($\downarrow$) & NIQE ($\downarrow$) & Q-Align ($\uparrow$) & CLIP-T ($\uparrow$) \\ \hline
Director3D~\cite{li2024director3d}      & 49.8   & 7.5    & 2.7  & \underline{23.5}       \\ 
LayerPano3D~\cite{yang2024layerpano3d}        & \underline{35.3}    & \underline{4.8}     & \underline{3.9} & 22.0       \\ 
\shortname (Ours)  & \textbf{34.6}    & \textbf{4.3}     & \textbf{4.2}   & \textbf{24.0}        \\        \hline
\end{tabular}
\vspace{3mm}
\caption{
Quantitative comparisons for text-to-world generation.
}
\label{tab:t2w}
\end{table}

%% file: sections/experiments/applications.tex
\subsection{Applications}
\shortname enables a wide range of practical applications given its three key advantages: 360° immersive experiences,  mesh export capability, and disentangled object modeling, as shown in~\figref{fig:application}.

\noindent\textbf{Virtual Reality.}
Our panoramic world proxy enables the generation of fully immersive 360° environments optimized for virtual reality deployment across contemporary VR platforms, such as Apple Vision Pro and Meta Quest. The comprehensive spatial coverage eliminates visual artifacts and boundary discontinuities, providing seamless omnidirectional browsing capabilities.

\noindent\textbf{Physical Simulation.}
The generated 3D worlds and separate 3D object representations support direct 3D mesh export, ensuring full compatibility with existing computer graphics pipelines for physical simulation. This enables seamless integration with physics engines for collision detection, rigid body dynamics, and fluid simulation.

\noindent\textbf{Game Development.}
The generated 3D worlds span diverse scenes with various aesthetic styles, including extraterrestrial landscapes, medieval architectural ruins, historical monuments, and futuristic urban environments. These worlds are exported as standard 3D mesh formats, enabling seamless integration with industry-standard game engines such as Unity and Unreal Engine for game development applications.

\noindent\textbf{Object Interaction.}
The disentangled object representations enable precise object-level manipulation and interaction within the generated 3D worlds. Users can perform precise 3D transformations, such as translation, rotation, and rescaling, on individual scene components while preserving the integrity of surrounding environmental elements. 

%% file: sections/related-work.tex
\section{Related Work}

\input{sections/related-work/panorama}

\input{sections/related-work/videoscene}

\input{sections/related-work/3dscene}

%% file: sections/related-work/panorama.tex
\noindent \textbf{Immersive Scene Image Generation.}
The $360^\circ$ panoramic image has emerged as a cornerstone for immersive virtual reality (VR) experiences. 
Recent advancements in latent diffusion models (LDMs)~\cite{rombach2022high,flux2024} have demonstrated remarkable capabilities in image synthesis.  Several studies including MVDiffusion~\cite{Tang2023mvdiffusion}, PanoDiff~\cite{wang2023360}, and DiffPano~\cite{ye2024diffpano} have successfully adapted diffusion models specifically for panorama generation. 
Subsequent efforts~\cite{wu2023panodiffusion,ye2024diffpano,zhang2024taming,zheng2025panorama} have focused on incorporating spatial priors, such as spherical consistency, depth to fine-tune pre-trained text-to-image (T2I) diffusion models on limited panoramic datasets. CubeDiff~\cite{kalischek2025cubediff} begins with a single cubemap face and synthesizes the remaining faces to reconstruct the complete panorama. 
In our work, we present a scalable panorama generation model for both text-conditioned and image-conditioned generation with a dedicated data curation pipeline.

%% file: sections/related-work/videoscene.tex
\noindent \textbf{Video-Based World Generation.}
Recent advances in video diffusion models (\eg Hunyuan-Video~\cite{kong2024hunyuanvideo}, CogVideo-X~\cite{yang2024cogvideox}, and Wan-2.1~\cite{wan2025wan}) have significantly improved high-quality video generation. Building on these models' inherent world knowledge, many methods now incorporate 3D constraints (e.g., camera trajectories, 3D points) to produce 3D-consistent videos for dynamic scene generation~\cite{he2025cameractrl,agarwal2025cosmos,alhaija2025cosmos,huang2025voyager}.
For instance, CameraControl~\cite{he2025cameractrl} and Cosmos~\cite{agarwal2025cosmos,alhaija2025cosmos} encode camera poses as Plücker coordinates, integrating them with latent embeddings to ensure camera-consistent video sequences. To achieve finer control, Streetscapes~\cite{deng2024streetscapes} employs multi-frame layout conditioning and autoregressive synthesis for long-range scene coherence. Meanwhile, Voyager~\cite{huang2025voyager} and Wu et al.~\cite{wu2025video} leverage explicit 3D scene points to enhance spatial consistency in extended video generation.
In game development, methods like Genie~\cite{parkerholder2024genie2} and Matrix~\cite{feng2024matrix} achieve   interactive video generation with  keyboard  actions. Similarly, many works generate videos in ~\cite{hu2023gaia,wang2024driving,gao2023magicdrive,wang2024drivedreamer} from text prompts, BEV maps, bounding boxes, and driver actions to simulate autonomous driving scenarios. However, these video-based methods suffer from high rendering cost and long-range inconsistency due to the lack of 3D representation.

%% file: sections/related-work/3dscene.tex
\noindent \textbf{3D World Generation.}
Compared to videos, 3D scene assets offer superior compatibility with standard computer graphics pipelines and ensure stronger consistency.
Existing 3D world generation methods can be categorized into  procedural-based methods~\cite{musgrave1989synthesis,raistrick2024infinigen,zhou2024scenex} and learning-based methods~\cite{yang2024layerpano3d,chung2023luciddreamer,yu2024wonderjourney,yu2025wonderworld,gao2024cat3d,liu2024physics3d,sun2024dimensionx}.
Procedural generation techniques automate the creation of 3D scenes by leveraging predefined rules or constraints. Among these, rule-based methods~\cite{musgrave1989synthesis,raistrick2023infinite} employ explicit algorithms to directly generate scene geometry, laying the foundation for visual rendering. In contrast, optimization-based generation~\cite{deitke2022️,raistrick2024infinigen} frames scene synthesis as an optimization problem, using cost functions based on predefined constraints (\eg, physics or design principles).
A more recent paradigm, LLM-based generation (\eg, LayoutGPT~\cite{feng2023layoutgpt}, SceneX~\cite{zhou2024scenex}) enables users to define environments through natural language descriptions, offering greater flexibility and user control over scene design.
Learning-based approaches focus on reconstructing 3D scenes from visual inputs. These methods typically fine-tune existing diffusion models using 3D-consistent data, enabling the generation of dense multi-view images from sparse known viewpoints. They then employ per-scene 3D/4D Gaussian Splatting optimization to reconstruct the full scene.
 Dimension-X~\cite{sun2024dimensionx} uses video diffusion models to train separate LoRA models, decoupling spatial and temporal dimensions for multi-view video generation.
Some works~\cite{chung2023luciddreamer,yu2024wonderjourney,yu2025wonderworld} generate novel-view images via progressive inpainting and world updating with depth as guidance.
Some methods \cite{schwarz2025recipe,yang2024layerpano3d,huang2025scene4u} approach 3D scene reconstruction by first inpainting occluded regions of a panorama, then lifting 2D views to 3D Gaussian Splatting (3DGS) through scene-specific optimization. Specifically, Layerpano3D~\cite{yang2024layerpano3d} proposed to disentangle different components in a panorama with depth clustering for 3DGS reconstruction, highlighting the effectiveness of layered representation. In contrast, our work targets generating layered mesh assets that can be directly plugged into existing computer graphics workflows.

%% file: sections/conclusion.tex
\section{Conclusion}
In this report, we introduced \shortname, a novel framework for generating immersive, explorable, and interactive 3D worlds from both text and image inputs. We leverage a semantically layered 3D mesh representation with a panoramic world proxy to create diverse and 3D-consistent worlds with disentangled objects for enhanced interactivity.  Extensive experiments demonstrate that \shortname achieves state-of-the-art performance for both text-based and image-based 3D world generation. The key features of our method---360° immersive experiences, mesh export capabilities, and disentangled object representations---enables a wide range of applications in virtual reality, physical simulation, and game development. We believe that \shortname represents a significant step forward in world-level 3D content creation and will serve as a valuable baseline for future research in this exciting and rapidly evolving field.
\\

\section*{Contributors}

\input{sections/contributors}

%% file: sections/contributors.tex
\begin{itemize}[leftmargin=0.25cm]
  \item \textbf{Project Sponsors:} Jie Jiang, Linus, Yuhong Liu, Di Wang, Tian Liu, Peng Chen
    \item \textbf{Project Leaders:}  Chunchao Guo, Tengfei Wang 
    \item \textbf{Core Contributors:}
Tengfei Wang, Zhenwei Wang,  Yuhao Liu, Junta Wu, Zixiao Gu, Haoyuan Wang, Xuhui Zuo, Tianyu Huang,  Wenhuan Li

    \item \textbf{Contributors:}   
    \begin{itemize}[leftmargin=0.5cm]
    \item \textbf{Engineering:} Sheng Zhang, Yihang Lian, Sicong Liu, Puhua Jiang, Xianghui Yang, Minghui Chen, Zhan Li, Wangchen Qin, Lei Wang, Yifu Sun, Lin Niu, Xiang Yuan, Xiaofeng Yang, Yingping He, Jie Xiao, Yangyu Tao, Jianchen Zhu, Jinbao Xue, Kai Liu, Chongqing Zhao, Xinming Wu
    \item \textbf{Data:} Lifu Wang, Jihong Zhang, Meng Chen, Liang Dong, Yiwen Jia, Chao Zhang, Yonghao Tan, Hao Zhang, Zheng Ye, Peng He, Runzhou Wu
   \item \textbf{Art Designer:} Yulin Tsai,  Dongyuan Guo, Yixuan Tang, Xinyue Mao, Jiaao Yu, Junlin Yu 
   
    \end{itemize}
\end{itemize}